\documentclass[a4paper,num-refs]{oup-contemporary}

\journal{general}

\usepackage{graphicx}
\usepackage{siunitx}
\usepackage{subfig}
\usepackage{tikz}

\title{A topological data analysis based classification method for multiple measurements}

\author[1]{Henri Riihimäki}
\author[2]{Wojciech Chachólski}
\author[3,4]{Jakob Theorell}
\author[3]{Jan Hillert}
\author[2,3\authfn{1}]{Ryan Ramanujam}

\affil[1]{Tampere University}
\affil[2]{KTH-The Royal Institute of Technology}
\affil[3]{Karolinska Institutet}
\affil[4]{University of Oxford}

\authnote{\authfn{1}ryan.ramanujam@ki.se; ryan@kth.se}

\papercat{Paper}

\runningauthor{Ramanujam et al.}

\jyear{2019}

\begin{document}

\begin{frontmatter}
\maketitle
\begin{abstract}
\textbf{Background}, Machine learning models for repeated measurements are limited. Using topological data analysis (TDA), we present a classifier for repeated measurements which samples from the data space and builds a network graph based on the data topology. When applying this to two case studies, accuracy exceeds alternative models with additional benefits such as reporting data subsets with high purity along with feature values. \textbf{Results}, For 300 examples of 3 tree species, the accuracy reached 80\% after 30 datapoints, which was improved to 90\% after increased sampling to 400 datapoints. Using data from 100 examples of each of 6 point processes, the classifier achieved 96.8\% accuracy. In both datasets, the TDA classifier outperformed an alternative model.  \textbf{Conclusions}, This algorithm and software can be beneficial for repeated measurement data common in biological sciences, as both an accurate classifier and a feature selection tool.
\end{abstract}

\end{frontmatter}


\section{Introduction}

Topological data analysis (TDA) is a recently emerging method for analyzing large-scale data using geometry and methods from algebraic topology \cite{CarlssonPointCloud,Oudot}. By considering topological features of multidimensional data together with various distance metrics imposed on the data, complex relationships within the data can be preserved and jointly considered. This often leads to better results than using standard analytical tools.

There have been several publications in biological research fields which have utilized TDA successfully. These include modeling RNA hairpin folding \cite{Singh2007}, Type-2 diabetes (T2D) subgrouping using clinical parameters \cite{LiT2D}, and classification of breast cancer tumors based on gene expression patterns \cite{Nicolau2011}.

Despite this, TDA software typically allows only singular measurements. That is, data is often input using a single measurement point per sample. Frequently in biological data collection, multiple measurement points are taken per sample. This may occur during sampling over some time interval, or repeated measures which are indicative of sampling from a distribution of events for each individual or sample. In this case, current methods are insufficient to classify these data accurately, since all measurement points are not considered together and in an informative manner.

To address these issues, we have developed a TDA based algorithm suitable for repeated measurements. This method also contains a classifier built on the network graph generated. This is accomplished using internal cross-validation using multiple bootstraps, and as a result the partitioning is robust against overfitting. The end result is a set of subgroupings of the relevant classes which can then be used as a starting point for further investigation into disease mechanisms.

We test this method on two unique data sets. The first is a simulation of six different point processes on a unit square. The second example is data from 3D modeling of various tree species, using laser scanning methods to determine characteristics of tree branches. These branches are then used as an input to the model. We demonstrate the accuracy of this method as compared to an support vector machine (SVM) based classifier as well as determining how the accuracy changes over given sampling rates, with the data available being large.

\section{Background}
\label{sec:background}

Topological data analysis arose from the mathematical field of algebraic topology. It was noted that in a set of data, some geometry might be present in the data distribution, or the data might reside on some topological structure such as a manifold. Even though these shapes are usually quite ideal, and obtaining data experimentally that conform to known mathematical shapes is highly improbable, topology looks at the global connectedness of the data set. Connections between data points correspond to relations in the data and topological methods give insight into this relational structure. Plainly, topological data analysis does not only look at the data points but also how they are globally related.

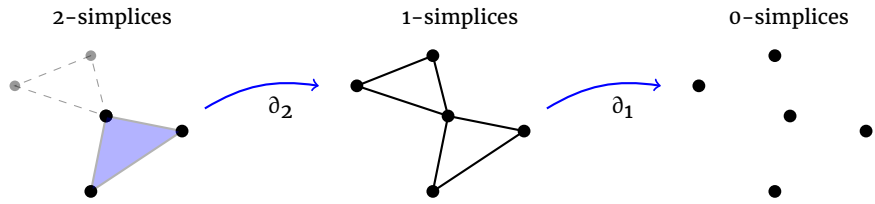
\begin{figure*}[ht]
	\centering
	\begin{tikzpicture}
	\tikzstyle{point}=[circle,thick,draw=black,fill=black,inner sep=0pt,minimum width=4pt,minimum height=4pt]
	\tikzstyle{shadowpoint}=[circle,thick,fill=black,inner sep=0pt,minimum width=4pt,minimum height=4pt,opacity=0.4]
	
	\node[right] at (-0.6,-0.1) {2-simplices};
	\node[right] at (4,-0.1) {1-simplices};
	\node[right] at (8.3,-0.1) {0-simplices};
	
	\node [shadowpoint] at (0,-0.6) {};
	\node [shadowpoint] at (-1,-1) {};
	\node [point] at (0.2,-1.4) {};
	\draw[dashed,opacity=0.4] (0,-0.6) -- (-1,-1) -- (0.2,-1.4) -- (0,-0.6);
	
	\node [point] at (1.2,-1.6) {};
	\node [point] at (0,-2.4) {};
	\node [point] at (0.2,-1.4) {};
	\path[draw=black,thick,fill=blue,opacity=0.3] (0.2,-1.4) -- (0,-2.4) -- (1.2,-1.6) -- (0.2,-1.4);
	
	\node [point] at (4.5,-0.6) {};
	\node [point] at (3.5,-1) {};
	\node [point] at (4.7,-1.4) {};
	\path[draw=black,thick] (4.5,-0.6) -- (3.5,-1) -- (4.7,-1.4) -- (4.5,-0.6);
	
	\node [point] at (5.7,-1.6) {};
	\node [point] at (4.5,-2.4) {};
	\path[draw=black,thick] (4.7,-1.4) -- (4.5,-2.4) -- (5.7,-1.6) --(4.7,-1.4);
	
	\node [point] at (9,-0.6) {};
	\node [point] at (8,-1) {};
	\node [point] at (9.2,-1.4) {};
	\node [point] at (10.2,-1.6) {};
	\node [point] at (9,-2.4) {}; 
	
	\draw[->,color=blue,thick] (1.5,-1.3) to [bend left=20] (3,-1);
	\draw[->,color=blue,thick] (6,-1.3) to [bend left=20] (7.5,-1);
	\node [below] at (2.5,-1.05) {\large${\partial_2}$}; 
	\node [below] at (7,-1.05) {\large${\partial_1}$};
	\end{tikzpicture}
	\caption{Five datapoints in two-dimensional space and the associated simplicial complex. The full simplicial complex is the purple 2-simplex and the three connected 1-simplices drawn here dashed. The boundary matrices are ${\partial_1}$ and ${\partial_2}$. Homology of this complex in degree 0 is 1 saying that there is one connected component. Homology in degree 1 is 1 indicating the loop created by the dashed 1-simplices.}
	\label{simp_comp_demo}
\end{figure*}

When data is distributed unevenly, for example as a point cloud in some metric space, geometric structures called simplicial complexes derived from the data can yield important information. One computationally simple way to construct a simplicial complex is the $r$-parameterized Vietoris-Rips construction. One data point is a 0-simplex, or a vertex, $k+1$ datapoints pairwise at most distance $r$ apart in the associated metric of data span a $k$-simplex at scale $r$. For example, two datapoints create a 2-simplex, or an edge. Simplices put together make simplicial complexes that can contain different geometric features of the data such as connected components, or clusters, and holes. Detecting holes in data has gathered interest for example in data base community \cite{BigHoles}. Homology is an algebraic method to measure the amount of geometric features of different degrees. Homology in degree zero counts the number of clusters, homology in degree 1 counts the number of 1-dimensional loops etc. Computing homology is effectively matrix computations with so called boundary matrices that contain information on how different simplices are connected to each other. See Figure \ref{simp_comp_demo} for an example of a simplicial complex and its homology.

Early success of TDA came in \cite{CarlssonImages} where it was discovered that the space of patches of natural images conform to a well-known geometric object. Another early advance came in \cite{Nicolau2011} where a TDA algorithm called "mapper" was used to find a new subgroup of breast cancer with excellent survival prognosis. Mapper models data as a network graph by refining standard clustering algorithms with topological ideas. Namely, global clustering of the data may be inefficient, especially when the data distance metric is not Euclidean. Instead data is partitioned according to some intervals. These intervals are created by using a filter function, meaning a function on data under which each point has exactly one value on some interval of real numbers. Then, local clustering is achieved based on those datapoints which map to the same interval. The clusters make nodes of the data network. Intervals are overlapping by some predefined amount. Clusters with non-empty intersection of points mapping to the overlap of two adjacent intervals are then joined by an edge. This construction creates a simplicial complex of clusters representing the structure of data under the chosen filtering function. This modification to standard clustering gives more insight into the global structure of data as explained above. There are publicly available mapper versions, such as "Python Mapper", which can be used to analyze data in this fashion \cite{PythonMapper}.

This paper builds upon this foundation by integrating a sampling procedure for the data, as well as adding a machine learning classifier which reports the unbiased accuracy of the underlying model. Important nodes of interest can be detected, which may yield important information about the data space, and relationships to the main outcome.



\section{Data Description}


Two datasets were employed for this study. The first was a simulation of six different point processes on the unit square. Point processes have gathered interest in TDA community as case studies, see for example \cite{ APF, LimitTheorems, HypothesisTesting}. Let $X \sim PD(k)$ denote that random variable $X$ follows probability disribution $PD$ with parameter $k$. In particular, $\text{Poisson}(\lambda)$ denotes the Poisson distribution with event rate $\lambda$.
\smallskip

\noindent
\textbf{Poisson}: \quad We first sampled number of events $N$, where $N \sim \text{Poisson}(\lambda)$. We then sampled $N$ points from a uniform distribution defined on the unit square $[0,1] \times [0,1]$. Here $\lambda=400$.
\smallskip

\noindent
\textbf{Normal}: \quad Again number of events $N$ was sampled from $\text{Poisson}(\lambda)$, $\lambda = 400$. We then created $N$ coordinate pairs $(x,y)$, where both $x$ and $y$ are sampled from normal distribution $N(\mu,\sigma^2)$ with mean $\mu$ and standard deviation $\sigma$. Here $\mu=0.5$ and $\sigma = 0.2$.
\smallskip

\noindent
\textbf{Matern}: \quad Poisson process as above was simulated with event rate $\kappa$. Obtained points represent parent points, or cluster centers, on the unit square. For each parent, number of child points $N$ was sampled from $\text{Poisson}(\mu)$. A disk of radius $r$ centered on each parent point was defined. Then for each parent the corresponding number of child points $N$ were placed on the disk. Child points were uniformly distributed on the disks. Note that parent points are not part of the actual data set. We set $\kappa$=80, $\mu$=5 and $r=0.1$.
\smallskip

\noindent
\textbf{Thomas}: \quad Thomas process is similar to Matern process except that instead of uniform distributions, child points were sampled from bivariate normal distributions defined on the disks. The distributions were centered on the parents and had diagonal covariance $\text{diag} (\sigma^2,\sigma^2)$. Here $\sigma=0.1$. 
\smallskip

\noindent
\textbf{Baddeley-Silverman}: \quad For this process the unit square was divided into equal size squares with side lengths $\frac{1}{28}$. Then for each tile number of points $N$ was sampled, $N \sim \text{Baddeley-Silverman}$. Baddeley-Silverman distribution is a discrete distribution defined on values $(0,1,10)$ with  probabilities  $(\frac{1}{10},\frac{8}{9},\frac{1}{90})$. For each tile, associated number of points $N$ were then uniformly distributed on the tile.
\smallskip

\noindent
\textbf{Iterated function system (IFS)}:\quad  We also generated point sets with an iterated function system. For this a discrete distribution is defined on values $(0,1,2,3,4)$ with corresponding probabilities 
$\left(\frac{1}{3},\frac{1}{6},\frac{1}{6},\frac{1}{6},\frac{1}{6}\right)$. We denote this distribution by IFS. Number of points $N$ was then sampled,  $N \sim \text{Poisson}(\lambda)$, $\lambda = 400$. Starting from an initial point $(x_0,y_0)$ on the unit square, $N$ new points are generated by the recursive formula $(x_n,y_n) = f_i(x_{n-1},y_{n-1}),$ where $n \in \{1,...,N\}$, $i \sim \text{IFS}$ and the functions $f_i$ are given as
\[f_0(y,x) = \left(\frac{x}{2},\frac{y}{2}\right), f_1(y,x)= \left(\frac{x}{2}+\frac{1}{2},\frac{y}{2}\right), 
f_2(y,x) = \left(\frac{x}{2},\frac{y}{2}+\frac{1}{2}\right)\] 
\[f_3(y,x) = \left(\left|\frac{x}{2}-1\right|,\frac{y}{2}\right), 
f_4(y,x)= \left(\frac{x}{2},\left|\frac{y}{2}-1\right|\right). 
\]

\begin{figure*}[ht]
	\includegraphics[width=1.0\textwidth, keepaspectratio]{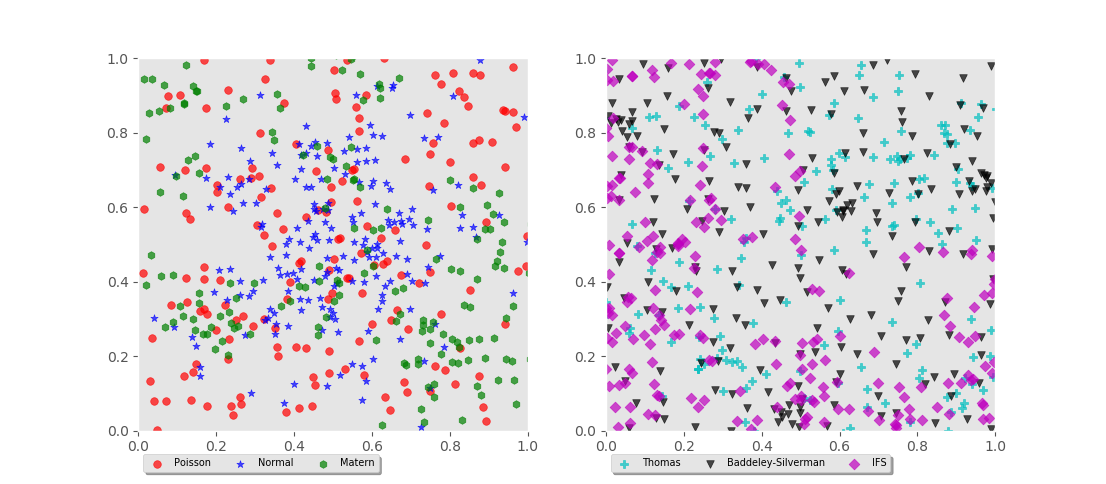}

	\caption{Example simulations of six different point processes used in the study.}
	\label{point_processes_fig}
\end{figure*}

\begin{figure}[ht]
	\centering
	\begin{tikzpicture}
	\draw[draw=black, ultra thick] plot [smooth,tension=1] coordinates{(0.75,1.1) (1.1,1.6) (1.4,2.3)};
	\draw[draw=black, ultra thick] plot [smooth,tension=1] coordinates{(2,1.75) (2.5,1.9) (2.9,2.2)};
	\draw[draw=black, ultra thick] plot [smooth] coordinates{(0.65,1) (1.2,1.2) (1.6,1.2) (1.9,1.25)};
	\draw[draw=black, ultra thick] plot [smooth] coordinates{(2.05,1.8) (2.2,2.3) (2.4,2.7)};
	\draw[draw=magenta, ultra thick] plot [smooth,tension=1] coordinates{(0,0) (0.75,1.1) (2,1.75) (3,3)};
	\end{tikzpicture}
	\caption{The purple main stem is the branch in our data sets. Purple part is the parent branch of the black child branches.}
	\label{branch_demo}
\end{figure}
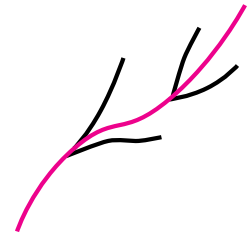

\begin{figure}[ht]
	\centering
	\includegraphics[width=0.55\columnwidth, keepaspectratio]{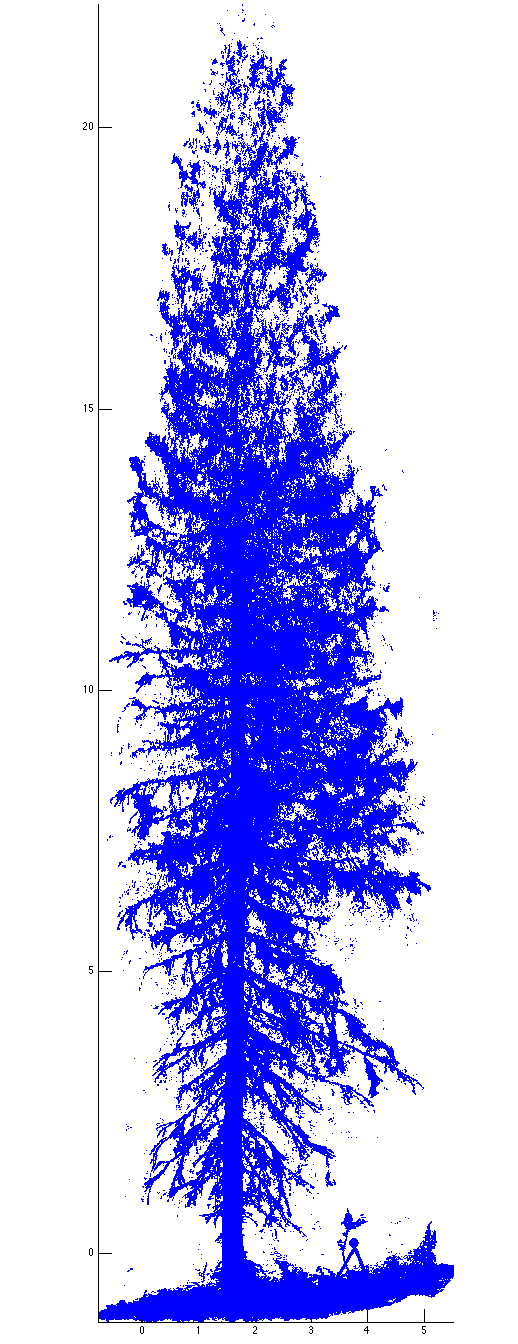}	
	\caption{Example of laser scanned spruce point cloud.}
	\label{spruce_cloud}
\end{figure}

\begin{figure}[ht]
	\centering
	\includegraphics[width=0.55\columnwidth, keepaspectratio]{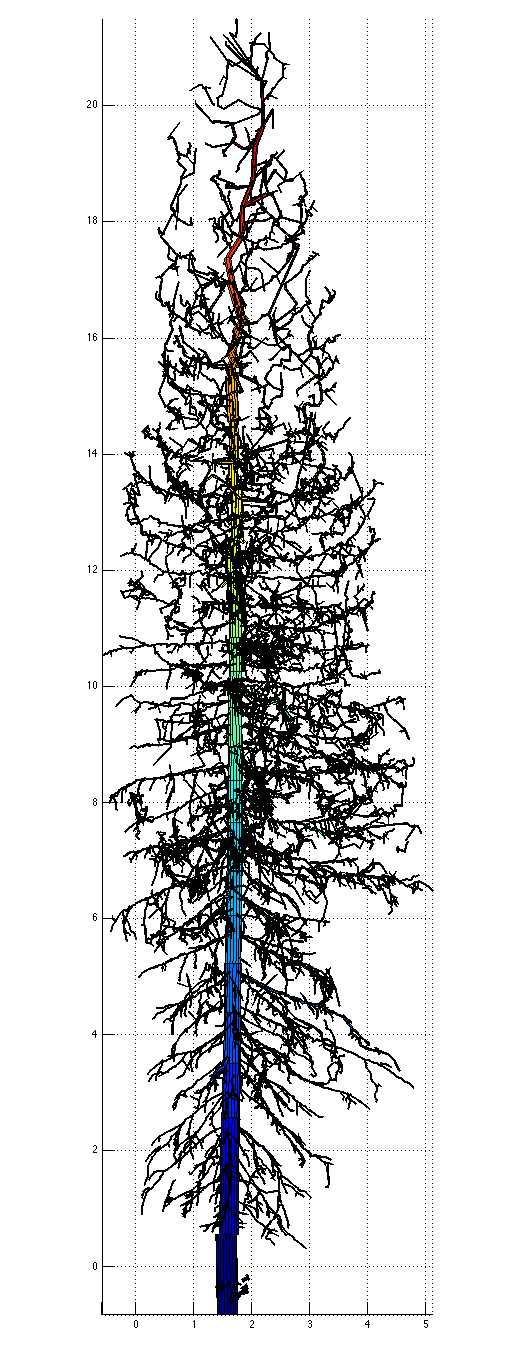}  	
	\caption{Example of QSM reconstructed model of the spruce point cloud in Figure \ref{spruce_cloud}.}
	\label{spruce_qsm}
\end{figure}

The second dataset came from Terrestrial Laser Scanning (TLS) of different tree species, representing a classification problem to correctly assign tree species from collected data. In this study we used data from Silver birch, Scots pine and Norway spruce. The scans were made in the location of Punkaharju in Finland. TLS produces point clouds of tree surfaces in 3D space. These point clouds can contain tens of millions of points and are not very useful for analysing tree data. Method of Quantitative Structural Modelling (QSM) for reconstructing tree models from TLS scans was developed in \cite{QSM}. This method reconstructs trees by fitting cylinders in the point clouds. Figures \ref{spruce_cloud} and \ref{spruce_qsm} show, respectively, examples of laser scanned point cloud of a Finnish spruce and its QSM reconstruction. Reconstructed models make it possible to obtain diverse data from trees. For example, lengths and volumes of individual branches are obtained by summing the lengths and volumes of the cylinders making up the branch. QSMs also contain the topological structure of trees as parent branch-child branch relations. For us branch means only the main stem excluding the child branches as shown in Figure \ref{branch_demo}.

Tree structures are ubiquitous in biological organisms. Some recent studies have applied topological data analysis methods on brain arteries \cite{Bendich2016} and neurons \cite{Kanari}. Biological tree structures are very naturally modelled as tree graphs in 3D space \cite{GodinCaraglio,Lamberton}. This however restricts the possibilities to obtain various data from the tree. Our approach is to view trees as point clouds of data and apply our topological analysis methods. As a multiple measurement case, we take one data point in a tree data set to be a branch of the tree with different features extracted from the QSM model. Specifically branch data point had features \{branch order (0 for trunk, 1 for branches originating from trunk etc.), branch length in meters, branch height above ground in meters, angle between branch and upward z-direction in radians\}. Trunk of the tree was excluded from the branch data.

\section{Methods}

The methodology used in this manuscript is an extension of published work in the TDA field combined with a machine learning approach. Due to this algorithm being created for repeated measurements, it is important to note that the term "sample" refers to one individual or one particular example of, i.e., a tree species, which itself contains many repeated measurement points. Each of these measurements is referred to as a "datapoint".

The algorithm presented here begins by randomly sampling each sample using some number of datapoints, less than or equal to the number of datapoints of the smallest sample included.  Thereafter a network graph was constructed, with nodes and edges representing small clusters of datapoints and connections between the clusters. This graph is built up by using the mapper algorithm as previously described. To summarize, mapper begins by simplifying the data space by using a one-dimensional filter function, by which each data point is assigned a value. Then, the range of values for this filter function are separated into overlapping intervals of some arbitrary length. Within these intervals, local clustering is conducted and guided by standard methods. The choice of linkage method here can be changed by the user. Nodes are then examined to see if datapoints occur in two different nodes mapping to the overlap of adjacent intervals. When this condition is true, an edge is drawn between the nodes. For purposes of this investigation edges are not necessary, and only the nodes contents themselves are used for analyses.

The next step of the algorithm is to add a machine learning on top of the structure. Nodes contain a number of datapoints from each of the samples based on geometry. This node information can be summarized in an $n$ by $m$ matrix where $n$ is the number of samples and $m$ is the number of nodes in the graph, and entries are the number of datapoints in a given node. These are then fed into a classifier, in this case a sparse logistic regression model was used for both binary and multi-class outcomes using the sklearn module SVC. After an unbiased classification accuracy was obtained, the last step was to rerun the entire data set, constructing a network graph from which feature selection could then occur based on the resultant classifier.

In order to avoid over fitting at any step, careful measures were taken. First, since this method samples from some data space, multiple samplings were conducted and the results were averaged to more accurately represent the sample distribution. Next, cross validation was conducted, as was running multiple classifiers in order to find the average results so that a particular data partitioning did not result in an over or under estimation of the classification accuracy. 

The general procedure was to first determine the sampling rates to use for each data set. These were [10,20,30,40,50,60,70,80,90,100,150,200,400] for the tree species and [10,20,30,40,50,60,70,80,90,100] for the point processes data. For each sampling rate in the point process data, 10 runs were conducted of the entire procedure, and the classification accuracy would be ultimately averaged for these runs. For tree species, this was increased to 100 runs due to larger variabiliy in the results. Within each run, a network graph was built upon which a cross-validated classifier could be built. The logistic regression model was created using 3-fold cross-validation, resulting in an out of sample prediction on each sample in the dataset.  The procedure of building a classifier was repeated for an alternative model, namely an SVM model with the optimal kernel. Via testing, this kernel was linear for the tree data and radial basis function (rbf) for the point processes. For alternative accuracy, the model was built on the sampled data as opposed to the network graph to provide the strongest possible alternative model. This also employed 3-fold cross-validation at the sample level, where a sample's out of sample prediction was based on the majority vote of its datapoints in the training SVM. The alternative models were also constructed 10 and 100 times to account for variability on cross-validation sample assignment.

Lastly, a single model to indicate variable importance was conducted on tree species data using the network graphing procedure and 400 sample points. Thereafter, information regarding the node size, average feature values for this node, and node purity were generated. Node purity is described as the proportion of datapoints in the node which belong to the largest class, such that the minimum can be 1/classes, and the maximum is 1. The average feature values for this node was determined by calculating the arithemtic mean of each feature for data in the node, providing a comparive mechanism to examine differences between nodes. This provides information regarding how values influence the outcome in a more complex manner than obtained with classical methods.

\section{Analyses}

The first set of analyses was to test the algorithm on the two aforementioned datasets. This was setup as a classification problem, wherein input data was used to predict the label. In both examples, the filter function used was the first principle component and the metric was Euclidean distance. The linkage selected was complete linkage. For point processes data, runs testing the TDA model as well as the alternative SVM model were conducted for sampling rates from [10..400], for both the full six point processes as well as for only normal and poisson point processes. The reason for the latter test was that the SVM appeared to have difficulty with the six class problem. The cross-validated accuracies are reported in Figure \ref{two_six_pp}. 

Using six point processes, the TDA accuracy was 60.7\% using 10 datapoints, and increased gradually to 96.8\% using 100 datapoints. The alternative SVM model began with 33.2\% using 10 datapoints, and remained at 33.3\% during sampling to 100 datapoints.

Using only the two mentioned point processes, the TDA classifier achieved an accuracy of 99.6\% after 30 datapoints, increasing to 99.975\% at 100 datapoints of sampling. The alternative SVM model achieved an accuracy of 99.1\% with 100 datapoints. 

\begin{figure*}[t!]  
	\centering
	\includegraphics[width=1.0\textwidth]{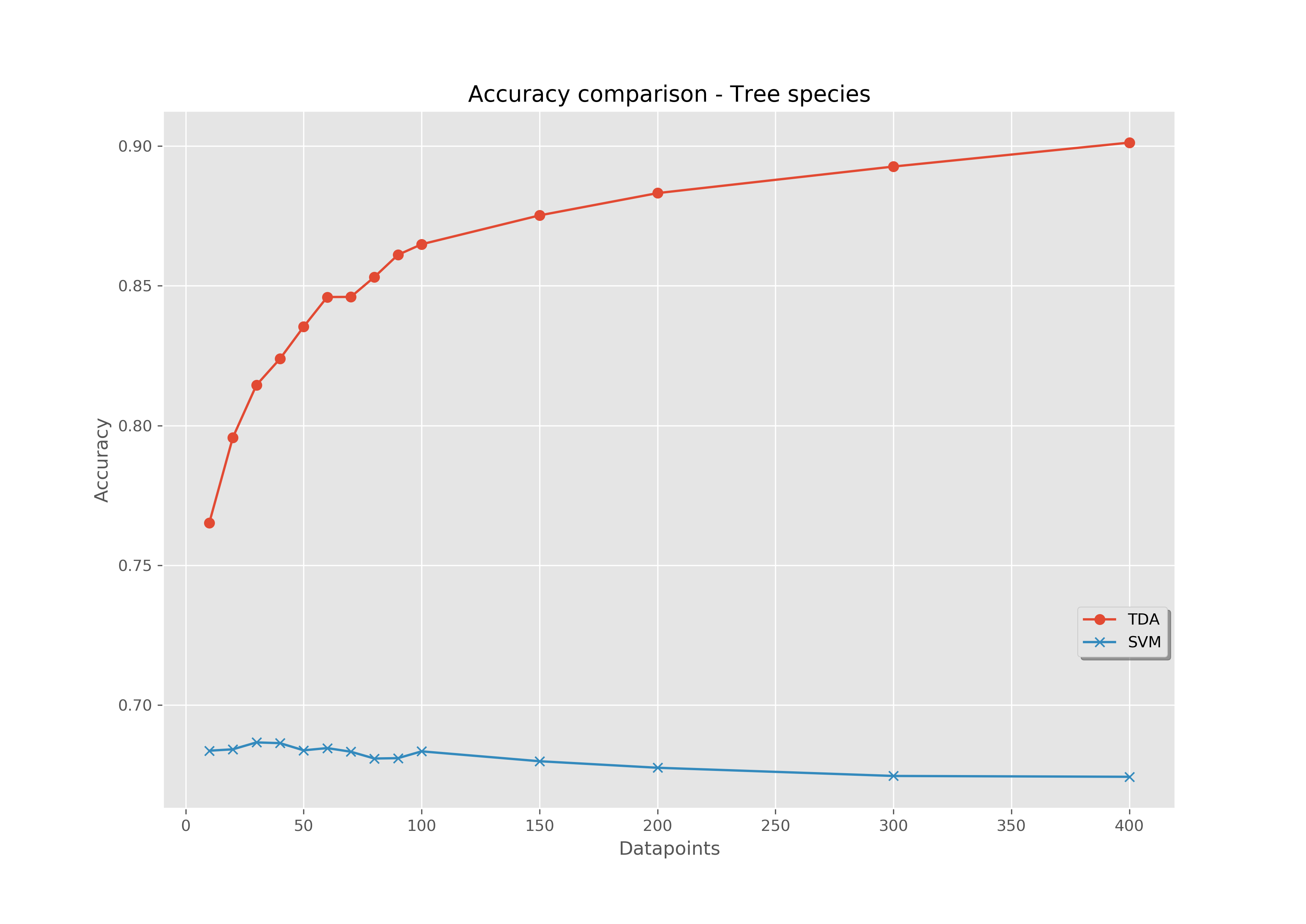}
	\caption{Cross-validated prediction accuracy when using a multi-measurement TDA classifier and compared with an SVM voting classifier. The dataset used were tree species, containing 100 examples of 3 classes. Datapoints are the number of points sampled per individual tree.
	}\label{tree_species}
\end{figure*}

The results of cross-validation for tree species is shown in Figure \ref{tree_species}. The TDA classifier had an accuracy of 76.5\% using 10 sampled datapoints, increasing in an asymptotic manner until 400 sampled datapoints and an accuracy of 90.1\%. The alternative model had an accuracy of 68.4\%, increasing to a maximum of 68.7\%  using 30 datapoints, thereafter reducing slightly.

The next analysis used node output generated from the software. Table \ref{node_table} presents the top nodes in the full data model using 400 datapoints, ordered by number of datapoints. The purity, i.e. largest class proportion, gives important information about the suitability of each node as a tool for subgrouping data into unique partitions. The additional columns signify the average values of each feature for the given node. Similarly, this data can be used to find nodes with certain characteristics which are present in the data which inform the class membership. Ideally, this data can be used on the original dataset to better understand data partitioning and subclusters of various classes.

\section{Discussion}

This paper presents a method to analyze data featuring repeated measurements, in order to obtain high classification accuracy as well as information regarding features in the data which are important for the outcome of interest. In biological data, repeated measurements are often obtained, for example when sampling the same individual over time, when large amounts of data are sampled at a high frequency, or when blood is sampled and a large number of measurement points are obtained. This algorithm builds on mapper (see above) and extends it into the realm of machine learning.

The sampling procedure demonstrates that often only a relatively small number of samples are required to adequately model the data space in question in order to get high classification accuracy. Our method is both highly accurate on these datasets when compared with other methods, and more informative than regression and more classical statistical techniques. Most importantly, it is able to determine which nodes are most responsible for the accuracy of the final model, such that important determinations about complex relationships in the data can be extracted.

\begin{figure*}[t!]  
	\centering
	\includegraphics[width=1.0\textwidth]{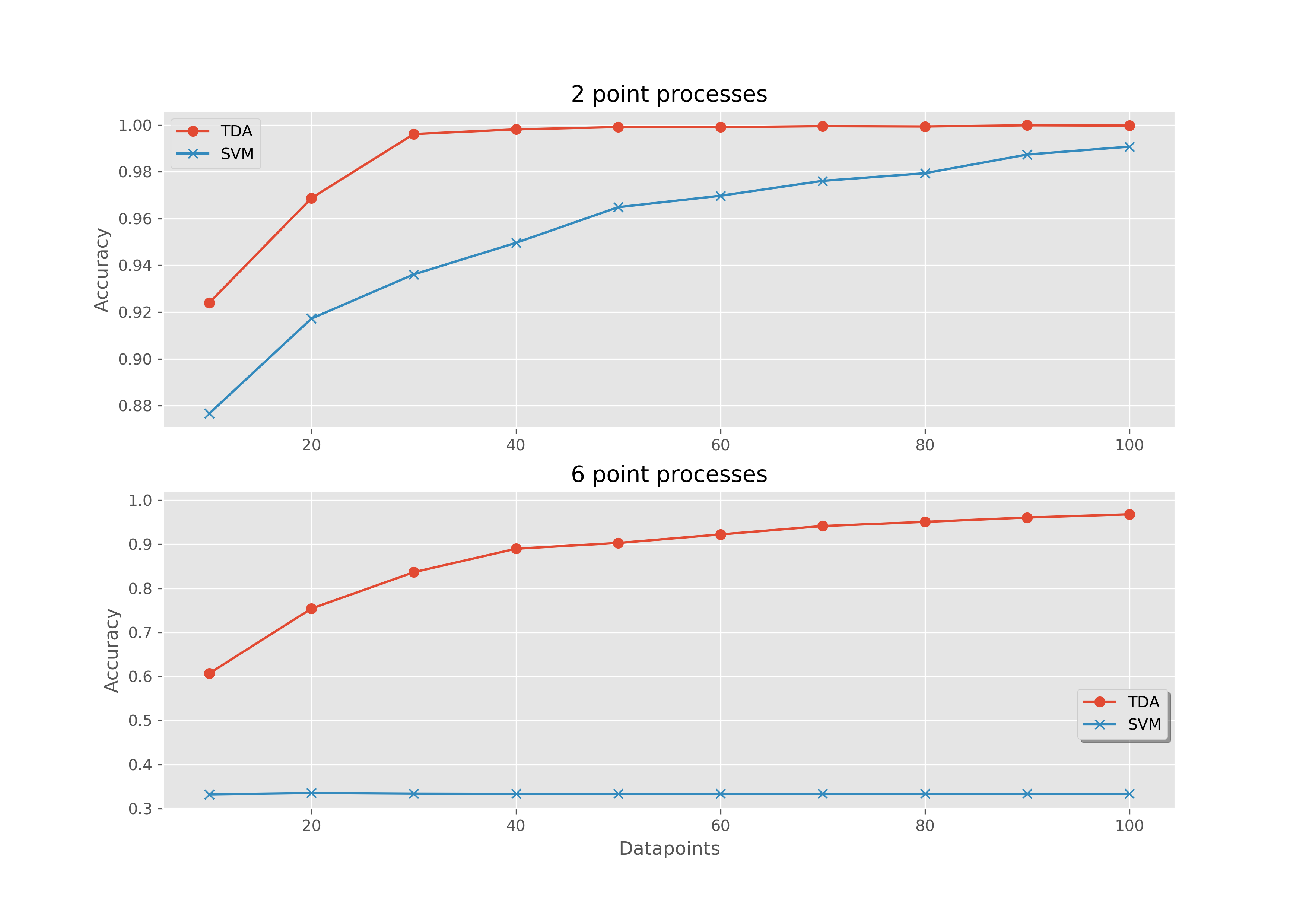}
	\caption{Cross-validated prediction accuracy when using a multi-measurement TDA classifier and compared with an SVM voting classifier. Datasets used were 2 classes of point processes (above) and 6 classes of point processes (below). Datapoints are the number of points sampled per example.
	}\label{two_six_pp}
\end{figure*}

When using the algorithms, there are a few caveats that need to be taken into consideration. First, the user must be avoid overfitting when the user tests a large number of filter functions and/or metrics, and selects the best one. Second, the results can be computationally intensive when the number of points sampled is high, due to runtime scaling to the order of $n^2$. This can be remedied partially by reducing the size of the intervals in the underlying algorithm, which could be automatically scaled for larger datasets. Lastly, since this algorithm uses internal cross validation the accuracy reported is based on a number of sub models which is equal to the number of cross validation intervals. The final model which determines node characteristics includes all samples, therefore may differ slightly from that from the internal data, which should not be confused with the unbiased estimate provided by the cross validated accuracy.

The cross validated accuracy of the TDA based classifiers exceeds the alternative SVM voting classifier in all tests and sampling rates presented. This was consistent despite using only a single metric and filter function for the TDA model, while selecting the best kernel for the SVM based on accuracy. 

An interesting note about classification accuracies is that with an increased number of classes, the presented algorithm maintains a high accuracy as shown in Figure \ref{two_six_pp}. For 6 point process, the alternate SVM classifier appears to maintain accuracy with two of the tree species, while confusing the other four species in a consistent basis, leading to the nearly constant 33\% cross-validation accuracy.This surprising phenomenon possibly reflects a large variation in the data which does not lead to data organization which is accurately partitioned by a hyperplane. Similarly, a potential explanation for the TDA classifier's high accuracy with more point process classes is that differences in datapoint location in multi-dimensional space could require tools to tease out clusters based on similar geometry.

\section{Potential implications}

The utility of this algorithm and implementation has broad applicability across the biological sciences as well as other fields. In particular, methods for obtaining repeated measurements classification models have been lacking, and our method fills a void in this manner. Furthermore, the ability to both partition data into its most useful components, and thereafter extract the features relevant for this partitioning, will allow researchers to identify which characteristics or variables in the data are most correlated with the outcomes.

Our algorithm and software can be employed by those who have repeated measurements data, and further extensions to this method can also be made. Also, the application of topological data analysis demonstrates a scenario wherein data geometry becomes useful, and depending on the data characteristics different metrics and filter functions can be applied. This demonstration of data analysis within the framework of machine learning and classification algorithms represents a novel utilization of TDA for common needs.

Additional development of methods using topological data analysis might result in further advances in classification techniques, and when combined with machine learning there is strong potential for these methods in the future.

\begin{table*}[b!]
	\caption{Top nodes by numbers of samples included, and average node feature values. Tree species data with 400 datapoints sampled from each data set.}\label{tab:example:wide}
	\begin{tabularx}{\linewidth}{S l l l l l l }
		\toprule
		{Node number} & {Datapoints} & {Purity} & {Branch order} & {Branch length} & {Branch height} & {Branch angle} \\
		\midrule
	17 &	23479 &	0.405 &	2.570 &	0.612 &	14.202 &	1.532 \\
	18 &	22887 &	0.604 &	2.627 &	0.655 &	17.436 &	1.511 \\
	14 &	22668 &	0.581 &	2.481 &	0.543 &	11.141 &	1.505 \\
	22 &	17634 &	0.766 &	2.710 &	0.650 &	20.322 &	1.494 \\
	10 &	13763 &	0.634 &	2.264 &	0.559 &	8.145 &	1.467 \\
	7 &	4481 &	0.536 &	1.641 &	0.758 &	5.076 &	1.393 \\
	6 &	3535 &	0.778 &	2.956 &	0.392 &	4.864 &	1.485 \\
	26 &	2614 &	0.600 &	1.947 &	0.860 &	23.349 &	1.453 \\
	2 &	2317 &	0.556 &	1.634 &	0.669 &	2.129 &	1.237 \\
	25 &	1821 &	0.784 &	3.484 &	0.439 &	23.041 &	1.543 \\
	9 &	1523 &	0.724 &	4.292 &	0.298 &	7.843 &	1.520 \\
	3 &	839 &	0.770 &	2.771 &	0.390 &	2.594 &	1.851 \\
	12 &	781 &	0.647 &	4.702 &	0.289 &	11.031 &	1.933 \\
		\bottomrule
	\end{tabularx}

	\begin{tablenotes}
		\item Purity is defined as the highest proportion of datapoints in the node that come from a single class.
		\item Branch order is the level of branching: 0 for trunk, 1 for branches originating from trunk etc.
		\item Branch length is the branch length in meters.
		\item Branch height is the height above ground in meters.
		\item Branch angle is defined as the angle between branch and upward z-direction in radians.
	\end{tablenotes}
    \label{node_table}
\end{table*}

\section{Availability of source code and requirements}

The source code for this algorithm is available as follows:
\begin{itemize}
\item Project name: Multiple measurements TDA classifier (mmTDA)
\item Project home page: ~\url{https://github.com/ryaram1/mmTDA}
\item Operating system(s): ~Platform independent
\item Programming language: ~Python
\item Other requirements: ~Python 3.0 or higher, numpy, pandas, scipy, sklearn, matplotlib, fastcluster
\item License: ~GNU General Public License v3.0 
\end{itemize}

\section{Acknowledgements}
The original TLS scans of the trees used in this study are the property of Raisa M\"{a}kip\"{a}\"{a}, raisa.makipaa@luke.fi, Natural Resources Institute Finland, Latokartanonkaari 9, FI-00790 Helsinki, FINLAND.

QSM models from the TLS scans were made by and are the property of Pasi Raumonen, pasi.raumonen@tut.fi, Tampere University of Technology, Korkeakoulunkatu 10, 33720 Tampere, FINLAND.

Ryan Ramanujam would like to acknowledge financial support by MultipleMS.

\subsection{Author's Contributions}
R. Ramanujam, H. Riihim\"{a}ki and W. Chachólski designed the study, with contributions from J. Theorell and J. Hillert. R. Ramanujam created the algorithm with W. Chachólski, and made the implementation. H. Riihim\"{a}ki and R. Ramanujam analyzed the data. The paper was jointly written by R. Ramanujam and H. Riihim\"{a}ki.





%
%
%
%
%
%
%
%
%
%

\bibliography{paper-refs}


\end{document}